%% file: sample-sigconf-authordraft.tex
\documentclass[sigconf]{acmart}
\usepackage{multirow}
\usepackage{booktabs}
\usepackage{multicol}
\usepackage{ifthen}
\usepackage{threeparttable}
\usepackage{enumitem}
\usepackage{subcaption}
\usepackage{tikz}
\usepackage[pages=some]{background}

\newboolean{anonymous}
\setboolean{anonymous}{false} %

\AtBeginDocument{%
  }

\ifthenelse{\boolean{anonymous}}{
  \setcopyright{acmlicensed}
  \copyrightyear{2026}
  \acmYear{2026}
  \acmDOI{XXXXXXX.XXXXXXX}
  \acmConference[MLCAD '26]{International Symposium on Machine Learning for CAD 2026}{September 7-9, 2026}{Jeju, South Korea}
  \widowpenalty=100     %
  \clubpenalty=100      %
}{
\copyrightyear{2026}
\acmYear{2026}
\setcopyright{cc}
\setcctype{by}
\acmConference[MLCAD '26]{2026 ACM/IEEE International Symposium on Machine Learning for CAD}{September 07--09, 2026}{Jeju Island, Republic of Korea}
\acmBooktitle{2026 ACM/IEEE International Symposium on Machine Learning for CAD (MLCAD '26), September 07--09, 2026, Jeju Island, Republic of Korea}
\acmDOI{10.1145/3831599.3840321}
\acmISBN{979-8-4007-2878-5/2026/09}
}

\begin{document}

\title{LOADBench: A Large Open Analog Dataset and Benchmark for Machine Learning in IC Design}

\ifthenelse{\boolean{anonymous}}{
  \author{Anonymous Authors}
  \affiliation{%
    \institution{Anonymous Institution(s)}
    \country{}}
  \renewcommand{\shortauthors}{Anonymous}
}{
  \author{Markus Leibl}
  \authornote{Both authors contributed equally to this research.}
  \affiliation{%
    \institution{TUM / Chair of Electronic Design Automation}
    \country{}
  }

  \author{Filipe Azevedo}
  \authornotemark[1]
  \affiliation{%
\institution{Instituto de Telecomunicações (IT) / IST-UL}
    \country{}
    }

  \author{Stefan Uhlich}
  \affiliation{%
    \institution{SonyAI}
    \country{}
  }

  \author{Andrea Bonetti}
  \affiliation{%
   \institution{SonyAI}
   \country{}
   }

  \author{Arun Venkitaraman}
  \affiliation{%
    \institution{SonyAI}
    \country{}
    }

  \author{Chia-Yu Hsieh}
  \affiliation{%
    \institution{SonyAI}
    \country{}
    }

  \author{Phuoc Pham}
  \affiliation{%
	\institution{TUM / Chair of DA}
	\country{}
    }

  \author{Lorenzo Servadei}
  \affiliation{%
    \institution{SonyAI, TUM}
    \country{}
    }

  \author{Helmut Graeb}
  \affiliation{%
    \institution{TUM}
    \country{}
    }

  \author{Ricardo Martins}
	\affiliation{%
	\institution{IT / IST-UL}
	\country{}
	}

  \renewcommand{\shortauthors}{Leibl, Azevedo et al.}
}

\begin{abstract}
Machine learning (ML) for analog IC design is constrained by limited open data and the absence of standardized evaluation: results reported in different papers are typically obtained on bespoke, often closed, datasets and with ad-hoc metrics, leaving comparisons across methods unreliable. We address both gaps with a large open-source dataset and a suite of standardized benchmarks. The dataset contains more than 100 operational amplifier (opamp) topologies (1\nobreakdash-, 2\nobreakdash-, and 3-stage) in the SkyWater130 PDK, each shipped with structural and functional annotations, topology-specific sizing constraints (symmetry and sizing rules), an Ngspice testbench, and a multi-objective NSGA-II sizing run producing around $10^3$ Pareto-optimal and $10^6$ intermediate sizing-performance pairs per topology. On top of this data we define six benchmark tasks with precise protocols: inverse sizing, topology selection, sub-circuit classification, structural circuit generation, Pareto-membership prediction, and multi-objective optimization. Thus, the data presented here constitutes the most comprehensive dataset for ML development in over four decades of analog IC design automation. The dataset, ground-truth annotations, and reference implementations of the metric-defining routines are released under an open license, enabling reproducible and directly comparable evaluation across heterogeneous analog design automation methods.
\end{abstract}



\maketitle
\backgroundsetup{opacity=1, scale=1, angle=0, contents={
		\begin{tikzpicture}[remember picture, overlay]
			\node[anchor=north east, inner xsep=50pt, inner ysep=10pt] at (current page.north east) {
				\href{https://www.acm.org/publications/policies/artifact-review-and-badging-current}{
					\includegraphics[width=50pt]{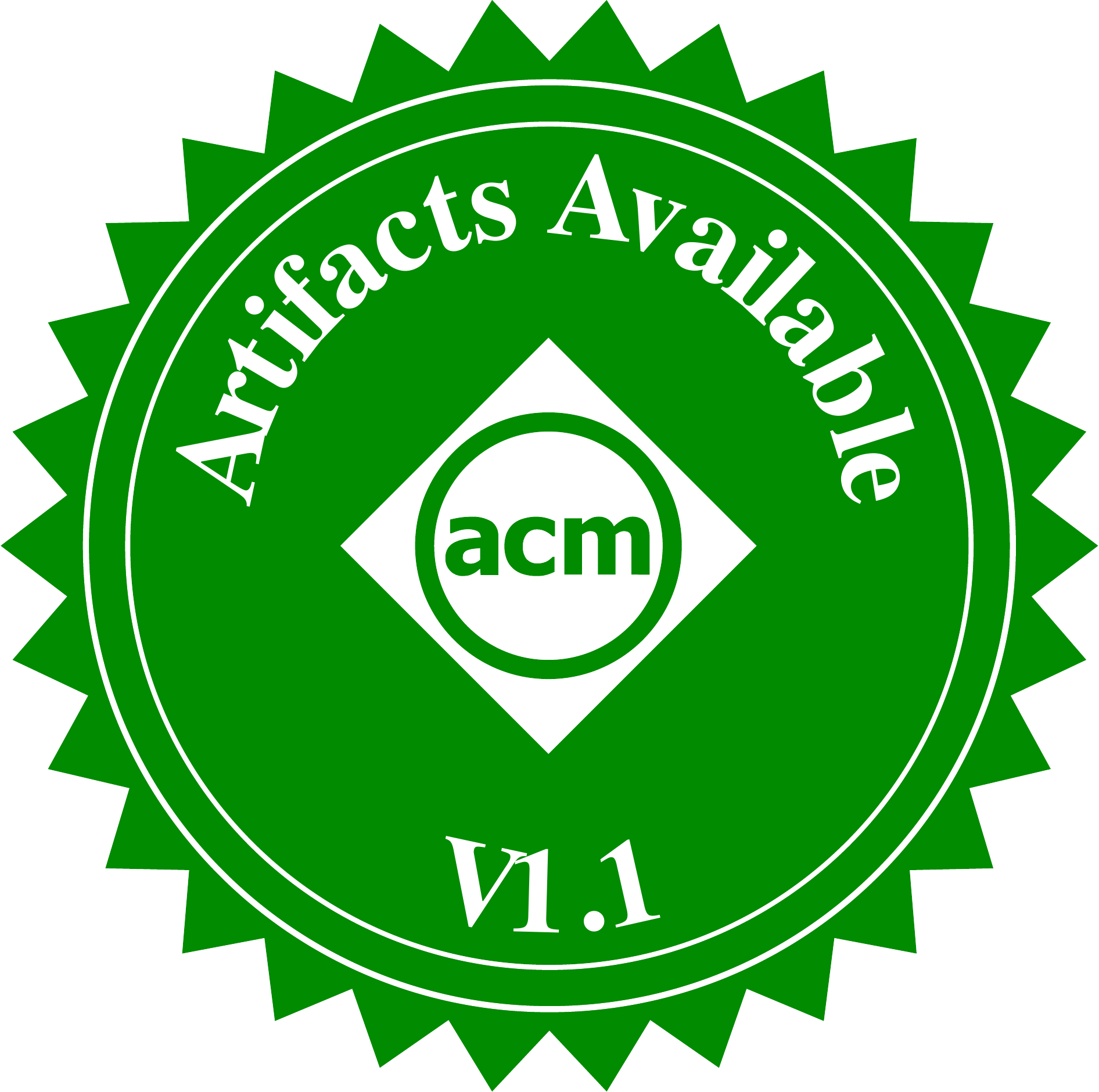}
					\includegraphics[width=50pt]{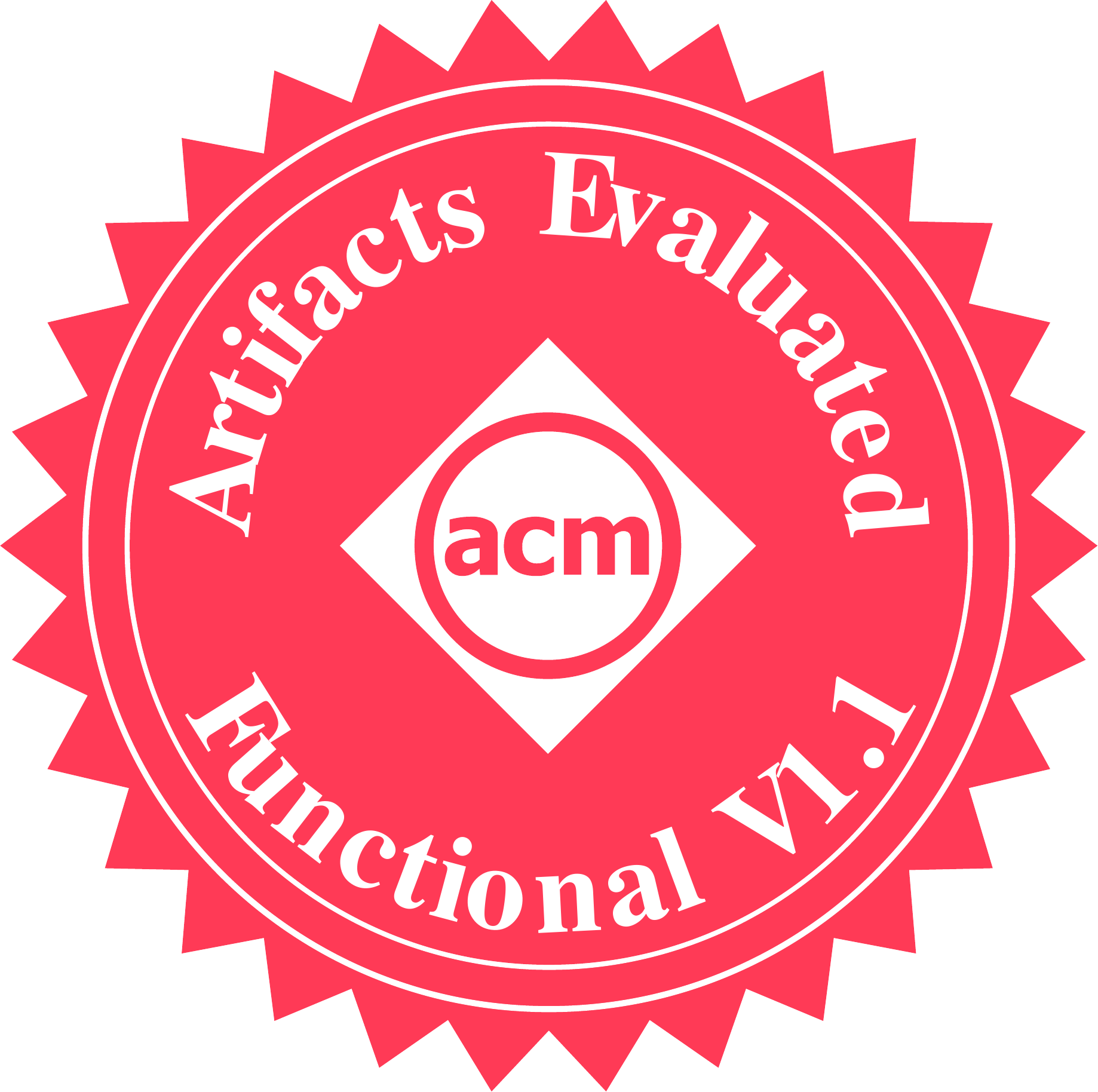}
			}};
		\end{tikzpicture}
}}
\BgThispage

\section{Introduction}\label{sec:introduction}

\input{introduction_v2.tex}

\section{Dataset Overview}\label{sec:dataset}
\input{dataset.tex}

\section{Proposed Benchmarks and Measurements}\label{sec:benchmarks}

\input{benchmarks.tex}

\begin{table*}[!tb]
	\centering
	\caption{Top-$k$ topology selection accuracy and mean first-hit rank (MFHR) for
		different metric-query settings.}
	\label{tab:selection_accuracy}
	\small
	\setlength{\tabcolsep}{3.2pt}
	{\footnotesize \begin{tabular}{llccccccccc}
		\toprule
		\multirow{2}{*}{Method}
		& \multirow{2}{*}{Query metrics}
		& \multicolumn{3}{c}{\textbf{Train}}
		& \multicolumn{3}{c}{\textbf{Validation}}
		& \multicolumn{3}{c}{\textbf{Test}} \\
		\cmidrule(lr){3-5}
		\cmidrule(lr){6-8}
		\cmidrule(lr){9-11}
		& & Top-1 & Top-3 & MFHR
		& Top-1 & Top-3 & MFHR
		& Top-1 & Top-3 & MFHR \\
		\midrule
		\textbf{Random baseline}
		& DC gain only
		& 80.4\% & 95.4\% & 2.07
		& 81.5\% & 94.3\% & 1.42
		& 75.5\% & 92.5\% & 1.61 \\
		& DC gain + F\textsubscript{t}
		& 53.2\% & 82.6\% & 3.62
		& 57.3\% & 84.8\% & 2.18
		& 56.3\% & 81.9\% & 2.43 \\
		& All active metrics
		& 3.2\% & 8.5\% & 43.80
		& 8.0\% & 23.6\% & 7.22
		& 7.8\% & 22.9\% & 7.57 \\
		\midrule
		\textbf{Learned embedding}
		& DC gain only
		& 86.4\% & 88.9\% & 1.37
		& 92.1\% & 100.0\% & 1.08
		& 91.9\% & 100.0\% & 1.14 \\
		& DC gain + F\textsubscript{t}
		& 80.1\% & 94.9\% & 1.45
		& 68.6\% & 99.0\% & 1.43
		& 86.7\% & 92.6\% & 1.44 \\
		& All active metrics
		& 15.7\% & 35.0\% & 10.56
		& 44.2\% & 74.9\% & 2.69
		& 37.2\% & 78.0\% & 2.76 \\
		\bottomrule
	\end{tabular}}
\end{table*}

\section{Evaluation and Reference Implementation for Two Benchmark Tasks}

\input{evaluation.tex}
\begin{table}[!htbp]
	\centering
	\caption{Pooled inverse-sizing results for one-stage opamps. Each row reports per-performance MAPE, APE standard deviation, outlier rate, spec-satisfaction rate, and $R^2$.}
	\label{tab:inverse_sizing_pooled_1_stage}
	\small
	\setlength{\tabcolsep}{3.2pt}
	{\footnotesize \begin{tabular}{lccccc}
		\toprule
		& MAPE (\%) & Std. (\%) & Outl. (\%) & Spec.\,(\%) & $R^2$ \\
		\midrule
		DC Gain                      &    1.90 &    5.61 &    0.20 &   46.48 &   0.94 \\
		F\textsubscript{t}                           &   13.42 &   26.78 &    3.32 &   49.80 &   0.98 \\
		Power                        &    8.03 &   13.68 &    2.54 &   52.34 &   0.99 \\
		Phase Margin                 &    5.36 &   15.82 &    1.56 &   53.12 &   0.04 \\
		$\text{SR}_\text{neg}$       &   26.28 &   79.91 &    9.57 &   50.00 &   0.90 \\
		$\text{SR}_\text{pos}$       &   22.55 &   50.23 &    4.10 &   55.02 &   0.96 \\
		V$_{\text{offset}}$          &   40.34 &  117.92 &   16.41 &   46.68 &   0.92 \\
		CMRR                         &    6.04 &   11.73 &    0.78 &   51.76 &   0.76 \\
		PSRR                         &    4.13 &   11.11 &    0.59 &   46.48 &   0.68 \\
		$\text{ICMR}_\text{min}$     &    5.84 &   10.93 &    1.56 &   49.80 &   0.88 \\
		$\text{ICMR}_\text{max}$     &    1.65 &    4.41 &    0.39 &   55.08 &   0.87 \\
		$\text{V}_\text{out,min}$     &    1.70 &    5.08 &    0.00 &   48.83 &   0.93 \\
		$\text{V}_\text{out,max}$     &    0.68 &    1.68 &    0.00 &   55.08 &   0.96 \\
		Area                         &    6.08 &    7.17 &    0.39 &   63.09 &   0.96 \\
		\bottomrule
	\end{tabular}}
\end{table}

\section{Conclusion}\label{sec:conclusion}
\input{conclusion.tex}

\section*{Acknowledgements}\label{sec:acknowledgments}
This research is funded by FCT – Fundação para a Ciência e a
Tecnologia, I.P., through national funds and, when applicable cofunded European Union (EU) funds under the projects UIDB/50008/2025 (DOI 10.54499/UIDB/50008/2025) and ACTON (DOI 10.54499/2023.1 1981.PEX).

\bibliographystyle{ACM-Reference-Format}
\bibliography{sample-base}

\appendix
\onecolumn

\section{Example Topologies and Pareto Fronts}
\label{app:example_topologies}
\input{example_topologies.tex}

\section{Complete Data for Inverse Sizing Example}
The pooled metrics for the 2-stage and 3-stage opamps are shown in Tables \ref{tab:inverse_sizing_pooled}(a) and \ref{tab:inverse_sizing_pooled}(b). 
\label{app:example_inverse}
\input{appendix_inverse.tex}

\section{Embedding Visualization}
\label{app:embedding_visualizations}
\input{embedding_visualization.tex}

\clearpage
\twocolumn
\label{app:artifact}

\input{ae.tex}

\end{document}

%% file: introduction_v2.tex
In contrast to the well-automated digital design flow, analog IC design remains a manual, knowledge-intensive, and iterative discipline, relying on the expertise of skilled engineers and lacking automatic design flows~\cite{Martins2012}. The resulting simulation-bound workflows prolong design cycles even as demand for analog-intensive applications continues to grow.

Machine learning has emerged as a promising tool to address these bottlenecks, but evaluation of ML methods for analog IC design remains fragmented. Unlike ML-dominated fields where benchmarks such as ImageNet~\cite{ImageNet} and COCO~\cite{COCO} for computer vision, or more recently Humanity's Last Exam~\cite{phan2025humanity} for large language models, provide universally accepted baselines, the analog community has not converged on an analogous evaluation framework. Reported results across papers are obtained on bespoke datasets and with ad-hoc metrics, leaving the field without a common ground for rigorous comparison.

\noindent To address this gap, our work introduces:
\begin{itemize}[leftmargin=*, itemsep=0pt, topsep=0pt]
	\item An open-source dataset of more than 100 opamp topologies (1\nobreakdash-, 2\nobreakdash-, and 3-stage) in the SkyWater130 CMOS PDK, generated with the open-source tool FUBOCO~\cite{FUBOCO} and sized with the software AIDA~\cite{AIDA}, each shipped with an Ngspice \cite{ngspice42} testbench and a full NSGA-II sizing run yielding roughly $10^3$ Pareto-optimal and $10^6$ intermediate sizing-performance pairs;

	\item Per-topology annotations of structural and functional transistor properties together with symmetry and sizing-rule constraints, enabling use across a broad range of design automation tasks;

	\item A benchmark suite of six tasks (inverse sizing, topology selection, sub-circuit classification, structural circuit generation, Pareto-membership prediction, and multi-objective optimization) with precise evaluation protocols, enabling direct comparison across diverse ML methods;
	
	\item A simple reference implementation in Python for data retrieval and metric calculation.
\end{itemize}

\noindent The presented \textbf{benchmark suite \cite{LOADBenchArtifact}} enables comparable research activities in the field and constitutes, to the knowledge of the authors, the most comprehensive analog dataset for ML yet.

\subsection{Analog IC Design Automation State-of-the-Art}

Machine learning has been applied across the analog design flow. LLM-based methods have been explored for topology selection/sizing and code/netlist generation~\cite{Vijayaraghavan2024, AnalogCoder, Nguyen2025}. Related non-LLM generative approaches include RL-based and evolutionary circuit generation~\cite{uhlich2024graco,uhlich2025spicemixer}, as well as variational autoencoders~\cite{Poddar2024}. For circuit sizing, surrogate models such as artificial neural networks (ANNs)~\cite{Islamoglu2019, Budak2022, Vaz2022} and decision trees~\cite{Martins2024} replace expensive simulator calls within optimization loops, while~\cite{POFComputationPaper} targets the Pareto-optimal front (POF) directly. The inverse sizing problem, mapping target specifications back to parameters, has been tackled through direct one-shot prediction using regression~\cite{Lourenco2018, Beaulieu2023}, dense neural networks~\cite{leibl24optimizerfree}, or generative models such as LLMs~\cite{Nguyen2025} and diffusion models~\cite{Azevedo2025, FilipeMarkus, DiffCktLiu2025}. Reinforcement learning has also been explored as an optimization-style alternative~\cite{Costa2025, Taskiran2025}. Beyond the circuit level, graph neural networks (GNNs) have been applied to placement-quality prediction~\cite{Li2020}, circuit classification~\cite{Kunal2020GANA}, topology generation \cite{dong2023cktgnn} and symmetry-constraint identification~\cite{Kunal2020, Gao2021, Xu2024}.

\begin{table}[t]
	\centering
	\caption{Comparison of open analog IC design benchmarks. ``Open tools'' indicates a fully open-source workflow (PDK, simulator where applicable); ``Struct.'' indicates whether structural annotations are shipped. ``Sizings'' is the total number of sizing--performance pairs.}
	\label{tab:benchmark_comparison}
	\small
	\setlength{\tabcolsep}{3pt}
	{\footnotesize \begin{tabular}{lccccc}
		\toprule
		Benchmark & \# Top. & Sizings & Open tools & Struct. & Scope \\
		\midrule
		AnalogGym~\cite{AnalogGym}                                          & 30   & --                & part. & no  & any \\
		AICircuit~\cite{AICircuit, mehradfar2025supervisedlearninganalogrf} & 9    & $3\!\times\!10^5$ & no    & no  & any \\
		AMSBench~\cite{AMSBench}                                            & --   & --                & --    & no  & LLM \\
		AnalogCoder~\cite{AnalogCoder}                                      & 24   & --                & --    & no  & LLM \\
		\midrule
		\textbf{Ours}                                                       & $\mathbf{>100}$ & $\mathbf{\sim 10^8}$ & \textbf{yes} & \textbf{yes} & \textbf{any} \\
		\bottomrule
	\end{tabular}}
\end{table}

\subsection{Analog IC Design Benchmarks and Datasets State-of-the-Art}

ML-based approaches depend on large training sets, a requirement that is especially pronounced for modern generative models: the diffusion model of~\cite{DiffCktLiu2025} relies on more than 400k datapoints, and the LLM of~\cite{Vijayaraghavan2024} on more than 800k. However, in the analog domain, non-disclosure agreements (NDAs) around commercial process development kits (PDKs) restrict the sharing of circuit data, limiting dataset availability, reproducibility, and cross-paper comparison~\cite{Li2024}, leading many researchers to fall back on synthetic data of uncertain fidelity~\cite{Poddar2024}.

Open benchmarks and datasets have emerged in response, but each leaves parts of the design space uncovered (Table~\ref{tab:benchmark_comparison}). AnalogGym~\cite{AnalogGym} ships only two of its five circuit categories in an open-source PDK. AICircuit~\cite{AICircuit, mehradfar2025supervisedlearninganalogrf} is built on commercial tools, restricting comparison to predicted-performance metrics on a closed test set. AMSBench~\cite{AMSBench} and AnalogCoder~\cite{AnalogCoder} target LLM schematic perception and code generation respectively, and are therefore confined to LLM-based approaches. The datasets OCB \cite{dong2023cktgnn}, AnalogToBi \cite{lee_analogtobi} and the EDA-Opamp-Database \cite{leibl24opampdatabase}, although extensive in number of netlists, have no sizing or complete structural annotations, the former two being built on commercial tools as well. None combines an open-source workflow, broad topology coverage, a benchmark suite spanning multiple stages of the analog IC design flow, and metrics applicable across a broad range of ML methodologies relevant to analog IC design. Furthermore, only AICircuit provides precomputed sizing data, at modest scale and behind a commercial PDK; methods developed against the other benchmarks must still generate their own training set, recreating the data-hunger problem that motivated open benchmarks in the first place. Moreover, when evaluation depends on a fixed dataset as in inverse sizing or classification, leaving its creation to the user undermines cross-method comparability.

\subsection{Organization of the Paper}

The remainder of the paper is organized as follows. Sec.~\ref{sec:dataset} presents the contents and generation pipeline of the proposed dataset. Sec.~\ref{sec:benchmarks} defines the six benchmark tasks together with their evaluation protocols and metrics. Sec.~\ref{sec:inverse_sizing_example} and~\ref{sec:topo_search_example} report reference implementations for the inverse-sizing and topology-retrieval benchmarks, illustrating how the dataset and protocols are used in practice. Additional examples and evaluations are presented in the Appendix.

%% file: dataset.tex
\begin{table*}[t!]
	\centering
	\caption{Performances and constraints used during dataset generation. The Preferred column indicates whether larger or smaller values are preferred. The HV ref.\ column lists the reference value used for hypervolume computation in Sec.~\ref{sec:optimization}; units match the Unit column.}
	\label{tab:performance_metrics}
	\small
	\setlength{\tabcolsep}{4pt}
	\begin{threeparttable}
		{\footnotesize \begin{tabular}{llclcc}
			\toprule
			Metric & Symbol / name & Unit & Preferred\tnote{a} & HV ref. & Performance constraints \\
			\midrule
			Area                              & Area                     & $\mu\mathrm{m}^{2}$ & smaller$^\star$ & 8000      & $\mathrm{Area} \leq 8000$ \\
			Common-mode rejection ratio       & CMRR                     & dB                  & larger          & --        & $\mathrm{CMRR} \geq 40$ \\
			Low-frequency (DC) gain                           & $\text{A}_{\mathrm{v,DC}}$                & dB                  & larger$^\star$  & 0         & $\text{A}_{\mathrm{v,DC}} \geq 0$ \\
			Transit frequency          & F\textsubscript{t}                       & Hz                  & larger$^\star$  & 0         &    $\text{F\textsubscript{t}} \geq 1000$ \\
			Maximum input common-mode voltage & ICMR\textsubscript{max}  & V                   & larger$^\star$  & 1.0       & $\text{ICMR\textsubscript{max}} \geq 1.0$ \\
			Minimum input common-mode voltage & ICMR\textsubscript{min}  & V                   & smaller$^\star$ & 1.1       & $\text{ICMR\textsubscript{min}} \leq 1.1$ \\
			Power                             & P                        & W                   & smaller$^\star$ & $1$ & $\text{P} \leq 1$ \\
			Phase margin                      & PM              & deg                 & larger$^\star$  & 50        & $50 \leq \mathrm{PM} \leq 90$ \\
			Power-supply rejection ratio      & PSRR                     & dB                  & larger          & --        & $\mathrm{PSRR} \geq 40$ \\
			Negative slew rate                & SR\textsubscript{neg}    & V/s                 & smaller$^\star$ & 0         & $\mathrm{SR}_{\mathrm{neg}} \leq 0$ \\
			Positive slew rate                & SR\textsubscript{pos}    & V/s                 & larger$^\star$  & 0         & $\mathrm{SR}_{\mathrm{pos}} \geq 0$ \\
			Input offset voltage              & V\textsubscript{offset}                  & V                   & smaller$^\star$ & 0.1       & -- \\
			Maximum output voltage            & V\textsubscript{out,max}  & V                   & larger$^\star$  & 1.1       & $\text{V}_{\mathrm{out,max}} \geq 1.1$ \\
			Minimum output voltage            & V\textsubscript{out,min}  & V                   & smaller$^\star$ & 1.0       & $\text{V}_{\mathrm{out,min}} \leq 1.0$ \\
			\bottomrule
		\end{tabular}}
		\begin{tablenotes}
			\item[a] Metrics marked with $^\star$ were optimized during the NSGA-II run that produced the dataset.
		\end{tablenotes}
	\end{threeparttable}
\end{table*}

The core of the benchmark data consists of >100 netlists of 1\nobreakdash-, 2\nobreakdash-, and 3-stage opamps (see Appendix~\ref{app:example_topologies} for representative schematics and Pareto-front visualizations) each of which is bundled with the following items:
\begin{itemize}[leftmargin=*, itemsep=0pt, topsep=0pt]
	\item  an XML file detailing structural information of the circuit, i.e. current mirrors, differential pairs, etc.;
	\item  an XML file detailing some functional properties of the circuit, e.g. stages, load, bias and transconductance parts;
	\item topology-implied sizing constraints, including symmetry and the sizing-rule constraints (per-device inversion/saturation margins, matching, etc.) of \cite{GZEA01};
	\item a SPICE-testbench which was used to generate the sizing data and delivers
	\begin{itemize}
	\item performance extractions for the metrics in Table \ref{tab:performance_metrics};
	\item necessary measurements for structural constraints: e.g. inversion, saturation, matching, etc.;
	\end{itemize}
	\item complete set of sizing data for each netlist containing a Pareto optimal front of \textasciitilde$1000$ points and all intermediate points amounting to \textasciitilde$10^6$ points in total;
	\item a trimmed SKY130 library for the used MOSFETs to speed up library loading in Ngspice.
\end{itemize}

\subsection{Structural and Functional Blocks}\label{sec:dataset_blocks}
\begin{figure}[]
	\centering
	\includegraphics[width=.7\linewidth]{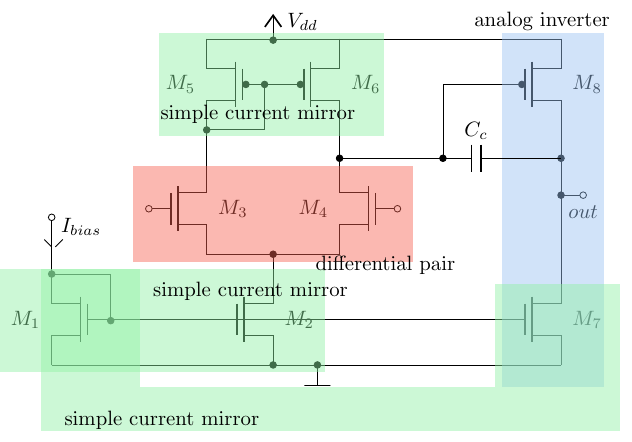}
	\setlength{\abovecaptionskip}{1ex}%
	\setlength{\belowcaptionskip}{0ex}%
	\caption{Example for recognized structural blocks.}
	\Description{}
	\label{fig:structrecDemo}
\end{figure}

\begin{figure}[]
	\centering
	\includegraphics[width=.7\linewidth]{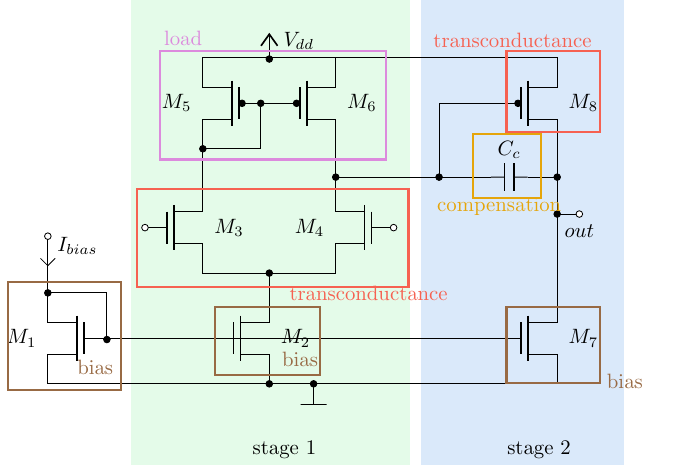}
	\setlength{\abovecaptionskip}{1ex}%
	\setlength{\belowcaptionskip}{0ex}%
	\caption{Example for recognized functional blocks.}
	\Description{}
	\label{fig:partDemo}
\end{figure}

The dataset comes with a structural and functional block classification which is described here. 
The algorithm is based on an extension of the original ACST software \cite{abel2022functional,abel2021acst}. It is close to the description in  \cite{abel2022functional}, but splits the classification into two parts. The first part comprises structural blocks such as various current mirrors and differential pairs. Those blocks are fully defined by their topological structure. The second part classifies circuit blocks into categories such as bias, transconductance, load or stages, which are not structurally defined, as there are usually several alternatives to their concrete implementation.
Examples of structural and functional block decomposition are shown in Figs.~\ref{fig:structrecDemo} and \ref{fig:partDemo} respectively.

\subsection{Testbenches}
The testbenches were written for the Ngspice simulator and include the DUT, i.e. the specific circuit we want to optimize and testbench circuitry for performance measure extraction of low-frequency gain, transit frequency, power, phase margin, slew rate, ICMR, V\textsubscript{out} range, PSRR, CMRR, V\textsubscript{offset}. They further include code to calculate those performances from the simulation data inside Ngspice's .control interpreter. The testbench includes a trimmed SKY130 library in order to reduce loading time, which is crucial for a fast simulation. 
Please note that the testbench generation focused on exploration of the performance space as well as execution speed. This is good for benchmarking machine learning tasks, but for concrete analog design the testbench should be optimized towards robustness instead. As a result, the dataset might contain points that are not good choices for an actual analog design, e.g. extreme points of the Pareto set or unwanted behavior in the time domain that was not caught in the short transient simulation.

\subsection{Sizing data}\label{sec:sizing_data}
The sizing data was created using the software AIDA \cite{AIDA}, which provides an NSGA-II optimizer that interacts with Ngspice for simulation-based multi-objective optimization. For each circuit, the optimizer was run for $1000$ generations with a population of $1024$. The optimization goals and the performance constraints applied during optimization are listed in Table~\ref{tab:performance_metrics}; performance constraints were set loosely to avoid hindering exploration, which sometimes results in design-irrelevant solutions in the dataset. In addition to the performance constraints, the optimization enforced symmetry constraints (using the per-netlist symmetry information shipped with the dataset) and structural sizing constraints following the sizing-rules method of~\cite{GZEA01}.

%% file: benchmarks.tex
To allow for consistent and meaningful comparison between different implementations, a set of benchmarks and performance indicators for each task enabled by the dataset is presented here.

\subsection{Circuit Generation}\label{sec:circuit_generation}

With more than 100 topologies, including universally accepted designs such as the Miller and the folded cascode opamp,
the LOADBench topology dataset $\mathcal{T}$ represents a set of canonical structures that a general-purpose opamp generator should be able to produce. We evaluate generators as a structural \emph{plausibility check}: does the generator output topologies that match these canonical designs? To remain applicable across generators that may target different PDKs or use simplified device models, the benchmark depends only on the structural representation of the netlist.

\textit{Match definition.}
Two netlists are a structural match iff there exists a bijection between their device sets that preserves device class (NMOS, PMOS, R, C, \ldots) and net connectivity, with terminal types preserved at each connection (gate, source, drain, bulk; drain and source treated as interchangeable for MOSFETs) and external ports preserved by role. Each reference netlist exposes a ground reference, a single supply rail, a bias-current input, a differential input pair, and a single-ended output, and the bijection must map each role to its counterpart in the candidate netlist. A Python reference implementation of this check is shipped alongside the dataset.

\textit{Conventions.}
The reference netlists use a single supply rail and ground, and include a complete bias network. Generators that use different supply or bias conventions can fall back to subgraph isomorphism: the submitter strips bias from their own outputs and checks whether the result is a subgraph of any reference in $\mathcal{T}$,
using the structural partitioning shipped with each reference to confirm that all amplification stages are present in the match. Such choices should be reported alongside the result.

\textit{Scope and metric.}
The benchmark applies in two modes: \emph{generation from scratch}, where the generator is run with no input and its outputs are matched against $\mathcal{T}$, and \emph{topology extension}, where the generator is given a starting netlist $T_0 \in \mathcal{T}$ and its derivative outputs are matched against $\mathcal{T} \setminus \{T_0\}$, testing incremental design moves such as adding a stage or introducing cascoding. Define the metric \textbf{Coverage}: the number of \emph{distinct} dataset topologies generated at least once, optionally normalized by $|\mathcal{T}|$ (or by $|\mathcal{T}|-1$ in extension mode). Generation time and hardware should be reported as diagnostic context. In addition to coverage as described here, we recommend reporting the diversity of the generated set by an approximate isomorphism check to discover duplicates.

\subsection{Topology Selection}
\label{sec:benchmarks:subsec:topo_selection}

Topology selection is an early and highly influential design decision in the
analog IC design flow. Identifying suitable topologies before detailed sizing can
substantially reduce the design search space and improve the efficiency of
automated design flows. Therefore, a topology-selection benchmark should evaluate
whether a method can retrieve circuit topologies that are capable of satisfying a
given target specification.

Topology selection can be formulated as a ranking problem: given a target
performance vector, the goal is to identify the circuit topologies whose sizing
solutions contain at least one point that Pareto-dominates, or satisfies, the
target specification. A method can therefore be evaluated by ranking all
candidate topologies according to their predicted suitability and checking
whether feasible topologies appear near the top of the ranking.

For each target performance vector, all candidate topologies are ranked according
to the score produced by the evaluated method. A topology is considered a
successful hit if at least one sizing solution in its Pareto-optimal set
satisfies the target specification, taking into account the optimization
direction of each performance metric. Hence, we propose top-$1$ and top-$3$
accuracy as primary metrics, which measure how often the first-ranked topology,
or one of the three highest-ranked topologies, contains a sizing solution that
achieves or exceeds the target performance vector.

In addition to top-$k$ accuracy, we propose the mean first-hit rank (MFHR),
which measures the average position of the first feasible topology in the ranked
list. For a query $q$, let $r_q$ denote the rank of the highest-ranked topology
that contains at least one sizing solution satisfying the target specification.
The metric is defined as
\begin{equation}
    \mathrm{MFHR} = \frac{1}{|\mathcal{Q}|}
    \sum_{q \in \mathcal{Q}} r_q ,
\end{equation}
where $\mathcal{Q}$ is the set of target performance queries. Lower values
indicate better retrieval performance, with an ideal value of $1$.

\subsection{Sub-Circuit Classification}\label{sec:subcircuit_classification}

The dataset ships per-netlist annotations from the ACST classifier (Sec.~\ref{sec:dataset_blocks}): structural blocks (current-mirror types, differential-pair types, \ldots) and functional blocks (stages, bias, transconductance, load). The benchmark evaluates whether a model can recover these annotations from the netlist alone: given a netlist, predict the set of structural and functional blocks present (multi-label). Evaluation uses a cross-circuit hold-out in which 10\% of the circuits are held out entirely, testing generalization to unseen topologies. Metrics are macro-averaged \textbf{accuracy}, \textbf{precision}, \textbf{recall}, and \textbf{F1}.

\subsection{Optimization}\label{sec:optimization}

For each circuit, the dataset records every sizing simulated by NSGA-II during POF construction (population $1024$, $1000$ generations, i.e.~$\sim10^6$ simulations per circuit), together with the resulting POF. This makes NSGA-II a natural reference for benchmarking other multi-objective optimizers on the same circuits, testbench, and PDK.

A candidate optimizer is run with the same budget as the reference dataset, using the same objectives, the performance constraints listed in Table~\ref{tab:performance_metrics}, and the symmetry and sizing-rule constraints shipped with each netlist (see Sec.~\ref{sec:sizing_data}). It is compared to NSGA-II by the \textbf{hypervolume ratio} $\mathrm{HV}_\mathrm{opt}/\mathrm{HV}_\mathrm{NSGA\text{-}II}$. Both hypervolumes are computed in the full 12-dimensional objective space against a fixed reference point chosen at the design-relevant worst on each axis (physical limit or constraint bound, listed in the ``HV ref.'' column of Table~\ref{tab:performance_metrics}). Exact 12D HV is intractable at this dimension, so we use the Bringmann--Friedrich FPRAS (pygmo's \texttt{bf\_fpras}, \cite{Biscani2020}) with approximation parameters $\varepsilon = 0.01$, $\delta = 0.01$, and a fixed random seed; a Python script applying this configuration is shipped alongside the dataset. A ratio of $1$ matches NSGA-II, above $1$ exceeds it. Submissions with limited compute may use a smaller budget $b$ instead: the dataset includes all intermediate POFs produced by NSGA-II, allowing a fair like-for-like comparison.

\subsection{Pareto Membership Prediction}\label{sec:pareto_optimal_points}

For each circuit, the dataset contains $\sim$ $10^3$ Pareto-optimal sizing-performance pairs together with $10^6$ dominated (intermediate) sizings produced during the NSGA-II run. This pairing supports a binary classification task: given a circuit topology and a candidate sizing $s$, predict whether $s$ lies on the recorded POF. A reliable predictor can support optimizers by identifying sizings unlikely to be Pareto-optimal or be used in conjunction with inverse sizing as described in Sec. \ref{sec:inverse_sizing}. Evaluation uses a per-circuit 80/10/10 random split of all sizings (POF and intermediate together), measuring how well a model trained on a dense sample of a circuit's design space recovers the POF on held-out sizings of the same circuit.

\textit{Metrics.}
With $\sim$ $10^3$ positives and $10^6$ negatives per circuit, the positive prior is below $10^{-3}$ and accuracy is uninformative (a constant-zero predictor exceeds $99.9\%$). Therefore we chose \textbf{ROC-AUC} as the metric, which is threshold- and imbalance-insensitive; \textbf{F1} at a tuned threshold is reported as a secondary number.

\subsection{Inverse Sizing}\label{sec:inverse_sizing}

Sizing assigns numerical values to a circuit's design parameters (transistor widths and lengths, capacitor and resistor values, bias currents) so that simulation yields a target performance. It is the inverse of simulation: simulation maps sizings to performances, sizing maps a desired performance back to a sizing that produces it. This benchmark targets methods that learn this map \emph{directly}, e.g., a neural network that emits a sizing in a single forward pass at inference time, rather than iterative optimizers that search the parameter space through repeated simulator calls.

\begin{figure}[h]
	\centering
	\includegraphics[width=.5\linewidth]{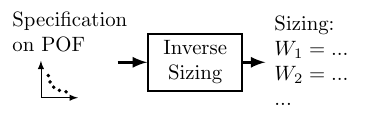}
	\setlength{\abovecaptionskip}{1ex}%
	\setlength{\belowcaptionskip}{0ex}%
	\caption{Benchmarked inverse sizing task.}
	\Description{}
	\label{fig:inversepic}
\end{figure}

We split the task into two subtasks: (i) projecting the user-supplied specification onto the circuit's Pareto-optimal front (POF), and (ii) mapping the resulting Pareto-optimal performance back to a sizing. Subtask (i) admits simple solutions in practice, polynomial fitting over the Pareto points works well~\cite{lourenco2019polynomial}. Subtask (ii) is more complicated, since POFs tend to exhibit richer structure in parameter space than in performance space, and is the focus of this benchmark (Fig.~\ref{fig:inversepic}).

To evaluate the sizing ability of different methods, we choose 15 different opamps to be part of the inverse sizing benchmark. These include five 1-stage, five 2-stage and five 3-stage opamps. An evaluation of more topologies might make the benchmark unattractive for methods that take a longer time to train, like diffusion models. Each opamp has around 1024 optimal sizing solutions which are split into training and test in a 90\%-10\% split. An approach should be trained on each opamp training set and then target the test set solutions belonging to the POF of each opamp of the benchmark and use the following measures to accurately evaluate their implementation:

\begin{itemize}[leftmargin=*, itemsep=0pt, topsep=0pt]
	\item The number of infeasible predicted sizing solutions (infinite values, NaNs, etc.);
	
	\item Mean absolute percentage error (MAPE) per performance;
	
	\item Standard deviation of MAPE per performance;
	
	\item Percentage of outliers that fail to satisfy a single specification by $30\%$ or more;
	
	\item Spec-satisfaction rate: the fraction of test samples whose predicted performance satisfies the queried target, with the direction taken from the ``Preferred'' column of Table~\ref{tab:performance_metrics};
	
	\item $R^2$ score;
	
	\item A Figure of Merit (FoM) aggregating prediction quality and simulation cost,
	\begin{equation}\label{eq:inverse_sizing_fom}
		\mathrm{FoM} \;=\; \frac{ \frac{1}{M} \sum_{m=1}^{M}\max(0,\,100-\mathrm{MAPE}_m)}{\bigl(1 + n_\mathrm{extra}/N_\mathrm{ref}\bigr)\,(1+o)},
	\end{equation}
	where $\mathrm{MAPE}_m$ is the per-metric MAPE in percent (clipped per metric so a single runaway metric cannot dominate the score); $o\in[0,2]$ sums the outlier and infeasibility fractions; $n_\mathrm{extra}\geq 0$ is the number of simulator calls beyond the provided dataset; and $N_\mathrm{ref}=1024\cdot 1000$ is the NSGA-II budget per POF (population $1024$, $1000$ generations). The cost factor $(1+n_\mathrm{extra}/N_\mathrm{ref})$ is floored at $1$, so using fewer simulations than provided gives no bonus, and replacing the provided POF with simulations from a different optimizer is out of scope. Under this convention $\mathrm{FoM}\in[0,100]$, with $100$ attained only by a perfect prediction without extra simulations;
	
	\item Training time and hardware used;
	
	\item In the case of generative models, the diversity of sampled solutions can be used as well.

\end{itemize}
To support readability, the per-spec metrics are to be reported as a pooled version for the set of 1\nobreakdash-, 2\nobreakdash-, and 3-stages.

%% file: evaluation.tex
\subsection{Topology Retrieval Example}\label{sec:topo_search_example}

Following the topology-selection benchmark introduced in
Sec.~\ref{sec:benchmarks:subsec:topo_selection}, we evaluate whether a learned
embedding model can rank circuit topologies from incomplete performance
specifications. This setting targets early-stage analog design, where only a few
requirements, such as gain or bandwidth, may be fixed, while the remaining
performance objectives are still open. To represent such partial specifications,
the metric encoder receives the normalized performance vector together with a
binary activity mask. Inactive metric entries are set to zero, and the mask
indicates which dimensions are actually part of the query. During training, we
randomly mask metric dimensions so that the model is exposed to both complete
and partial specifications.

We train a CLIP-style dual encoder~\cite{radford2021learning} that maps topology
graphs and Pareto-optimal performance vectors into a shared embedding space. The
topology tower is a 3-layer GCN with 128 hidden channels, batch
normalization, ReLU activations, dropout, global mean pooling, and a 2-layer
projection head to a 64-dimensional L2-normalized embedding. The metric tower is
an MLP that takes the metric vector concatenated with its activity mask and
produces an embedding of the same dimension. The model is optimized with a
symmetric contrastive loss, where Pareto-optimal performance vectors are treated
as positives for their corresponding topology.

For evaluation, each Pareto-optimal sizing is used as a query. Candidate
topologies are ranked by cosine similarity between the query embedding and the
topology embeddings. A retrieval is counted as correct if at least one
Pareto-optimal sizing of the retrieved topology satisfies all active query
metrics, respecting whether each metric is maximized or minimized. Importantly,
each query is evaluated against only the topologies belonging to the same
split, i.e., for the test split, both the query sizing and the candidate topologies
are unseen during training. Hence, the evaluation reflects generalization to
genuinely unknown circuits. We also
report an analytic random baseline under the same satisfaction criterion: for
each query, we count the number of satisfying topologies and compute the exact
probability that a uniformly random top-$k$ ranking contains at least one of
them. This baseline is high for weak partial specifications, since many
topologies can satisfy a single-metric target. For MFHR, we use the corresponding
expected first-hit rank under a uniformly random ranking.

Table~\ref{tab:selection_accuracy} reports top-$k$ accuracy and MFHR for the
learned embedding and the random baseline. For DC gain-only queries, the random
baseline is already strong, but the learned embedding reaches 91.9\% top-1 and
100.0\% top-3 accuracy with an MFHR of 1.14 on the test split. Adding F\textsubscript{t}
makes the query more selective: the random top-1 baseline drops to 56.3\% with
MFHR~2.43, while the learned embedding achieves 86.7\% top-1 and 92.6\% top-3
accuracy with MFHR~1.44. Retrieval using all active metrics is substantially
harder. In this setting, the learned embedding improves over the random baseline,
reaching 37.2\% top-1 and 78.0\% top-3 test accuracy with MFHR~2.76 compared
to the baseline MFHR of 7.57, but the absolute performance indicates that exact
multi-objective topology retrieval remains challenging and should be tackled in future work.

Additional UMAP visualizations of the learned topology--metric embedding space
for the query settings are provided in
Appendix~\ref{app:embedding_visualizations}.

\subsection{Inverse Sizing Example}\label{sec:inverse_sizing_example}
\begin{figure}[h]
	\centering
	\includegraphics[width=.9\linewidth]{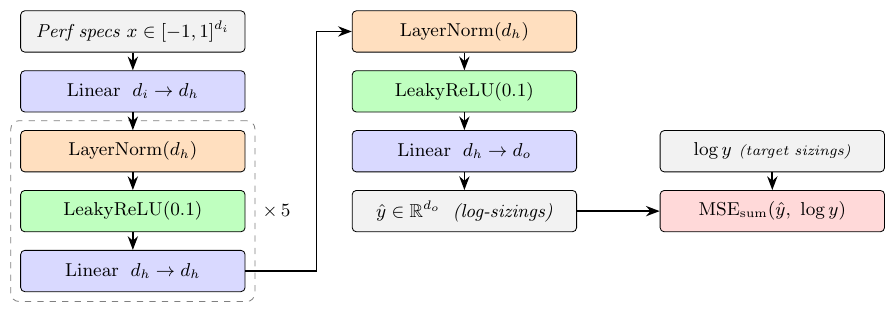}
	\setlength{\abovecaptionskip}{1ex}%
	\setlength{\belowcaptionskip}{0ex}%
	\caption{FCNN and loss for inverse sizing example.}
	\Description{}
	\label{fig:fcnn}
\end{figure}

We evaluate a fully-connected neural network (FCNN) reference implementation of the inverse sizing benchmark (Sec.  \ref{sec:inverse_sizing}).
The FCNN for a specific circuit topology takes the $12$ performance measures that were used as goals for the genetic optimization from Table \ref{tab:performance_metrics} (marked with a *) as input and outputs the complete sizing. 

The network architecture is depicted in Fig. \ref{fig:fcnn}. It uses the input dimension $d_i$ as the number of input performances, the hidden dimension $d_h = 48\times d_i$ and the output dimension $d_o$ as the number of individual sizing parameters. Performances are normalized to the interval $[-1,1]$. Sizings are normalized as follows: Widths are scaled by lengths, $\frac{W_j}{L_j}$, lengths are scaled to $\mu m$, capacitances are scaled to $10 pF$. All sizings are transformed to the log-space using the natural logarithm. We use PyTorch's NAdam optimizer with $\text{lr}=2\cdot10^{-3}$, $\text{weight\_decay}=10^{-5}$, $\text{eps}=10^{-8}$. Gradients are clipped with max\_norm=$1.0$ after every backward propagation. We employ PyTorch's \textit{ReduceLROnPlateau} scheduler with mode='min', factor=$0.5$, patience=$10$, $\text{min\_lr}=10^{-6}$. Early stopping with a window of $20$ is used on the validation set. After the optimization stops, a second optimization is started with the LBFGS optimizer, lr=$1$, line\_search\_fn='strong\_wolfe', max\_iter=$300$, tolerance\_grad=$10^{-7}$, tolerance\_change=$10^{-11}$, history\_size=$20$. No scheduler is used for this run, but early stopping is set to $3$. The evaluation for the test set of one-stage opamps as described in Sec. \ref{sec:inverse_sizing} can be seen in Table \ref{tab:inverse_sizing_pooled_1_stage}. For the other two sets, please see Appendix~\ref{app:example_inverse}, Table \ref{tab:inverse_sizing_pooled}. The aggregate FoM is $57.83$, while it is $69.86$, $69.80$ and $27.81$ for 1\nobreakdash-, 2\nobreakdash-, and 3-stage opamps respectively. Likewise, aggregate infeasible predictions are $2.73 \%$, $8.82\%$ and $61.88\%$. The average training time per circuit was 109 seconds, on a laptop equipped with an Intel i7-13700H CPU and an NVIDIA GeForce RTX 4060 Ti GPU.

%% file: conclusion.tex
We presented LOADBench, an open \textbf{dataset \cite{LOADBenchArtifact}},  of more than 100 annotated opamp topologies in the SkyWater130 PDK, together with six standardized benchmark tasks. By releasing the data, ground-truth annotations, and reference metric implementations under an open license, we provide a common basis for directly comparable evaluation of machine-learning methods for analog IC design. We hope LOADBench lowers the entry barrier for new methods and enables reproducible progress on a problem that has long suffered from fragmented, closed evaluation.

%% file: example_topologies.tex
\begin{figure}[h]
	\centering
	\begin{subfigure}[t]{0.24\linewidth}
		\centering
		\includegraphics[width=\linewidth]{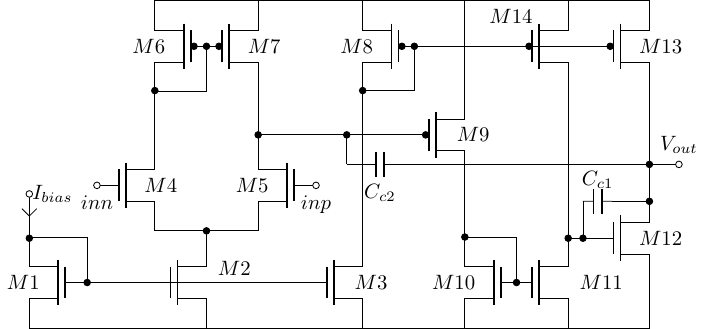}
		\caption{3-stage opamp with nested Miller compensation.}
		\label{fig:3stage}
	\end{subfigure}\hfill
	\begin{subfigure}[t]{0.24\linewidth}
		\centering
		\includegraphics[width=\linewidth]{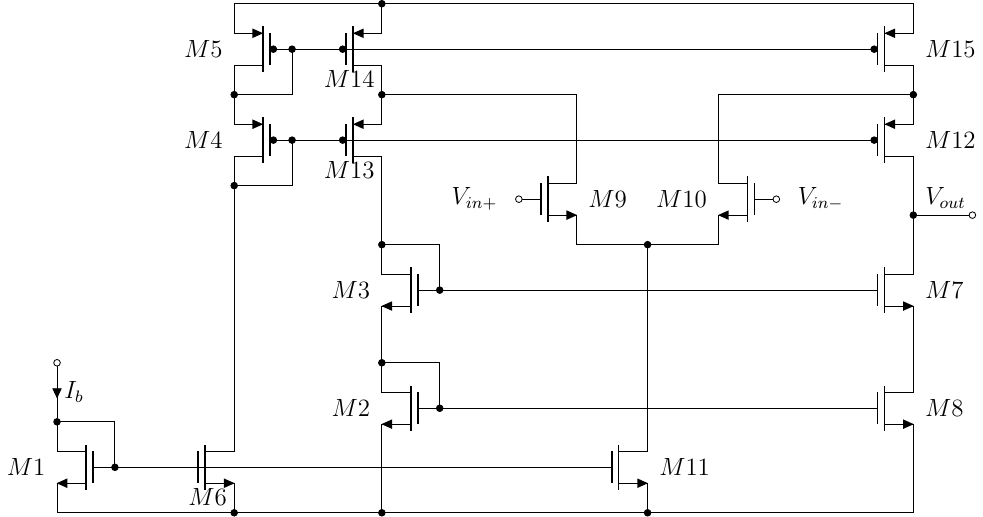}
		\caption{Folded-cascode opamp.}
		\label{fig:fcoa}
	\end{subfigure}\hfill
	\begin{subfigure}[t]{0.24\linewidth}
		\centering
		\includegraphics[width=.9\linewidth]{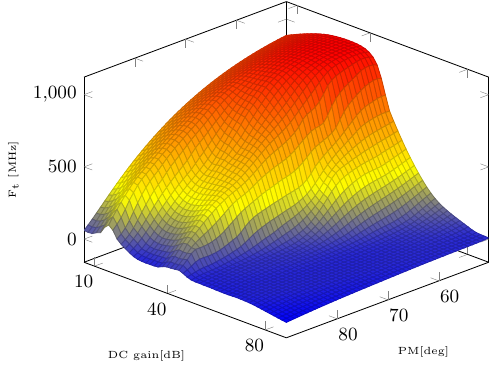}
		\caption{Projection onto the (phase margin, DC gain, F\textsubscript{t}) subspace.}
		\label{fig:pmgainftfront}
	\end{subfigure}\hfill
	\begin{subfigure}[t]{0.24\linewidth}
		\centering
		\includegraphics[width=.9\linewidth]{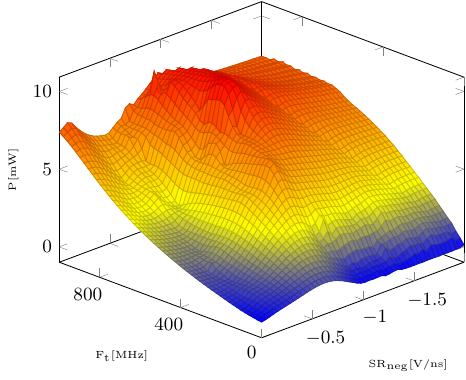}
		\caption{Projection onto the (F\textsubscript{t}, negative slew rate, power) subspace.}
		\label{fig:ftsrfront}
	\end{subfigure}
	\caption{Two example opamp topologies from the dataset and interpolated 3D projections of the POF of opamp \ref{fig:fcoa}.}
	\label{fig:example_opamps}
\end{figure}

As examples of the netlists in the dataset, Fig.~\ref{fig:example_opamps} shows two opamps and two 3D projections of the POF of the folded-cascode topology.

%% file: appendix_inverse.tex
\begin{table}[!htbp]
	\centering
	\caption{Pooled inverse-sizing results. Metrics and performances as in
		table~\ref{tab:inverse_sizing_pooled_1_stage}.}
	\label{tab:inverse_sizing_pooled}
	\setlength{\tabcolsep}{1.2pt}
	{\scriptsize
		\begin{tabular}{l*{14}{c}@{\hskip 10pt}*{14}{c}}
			\toprule
			& \multicolumn{14}{c}{(a) 2-stage opamps}
			& \multicolumn{14}{c}{(b) 3-stage opamps} \\
			\cmidrule(lr){2-15}\cmidrule(lr){16-29}
			& Gain & F\textsubscript{t} & P & PM & SR$_-$ & SR$_+$ & V$_{\text{off}}$ & CMRR & PSRR & $\text{ICMR}_-$ & $\text{ICMR}_+$ & $\text{V}_\text{o+}$ & $\text{V}_\text{o-}$ & Area
			& Gain & F\textsubscript{t} & P & PM & SR$_-$ & SR$_+$ & V$_{\text{off}}$ & CMRR & PSRR & $\text{ICMR}_-$ & $\text{ICMR}_+$ & $\text{V}_\text{o+}$ & $\text{V}_\text{o-}$ & Area \\
			\midrule
			MAPE (\%)   &  3.68 & 10.26 &  6.00 &     6.36 & 22.25 & 14.04 &  40.15 &  6.00 &  8.74 &  5.99 &  6.14 &  4.36 &  0.99 &  3.47 &  51.68 &  96.33 & 10.92 &   74.32 &  37.82 & 37.51 &     169.96 &  21.85 &  71.29 & 17.69 & 22.87 &  81.43 & 12.19 &  4.35 \\
			Std. (\%)   &  9.24 & 40.57 & 10.59 &    52.49 & 88.08 & 51.28 & 138.44 & 12.45 & 14.37 & 10.09 & 17.65 & 17.31 &  1.96 &  4.72 & 133.73 & 242.49 & 35.88 &  146.36 & 134.45 & 87.91 &     341.48 &  58.21 & 189.67 & 26.65 & 32.98 & 150.27 & 22.73 &  4.61 \\
			Outl. (\%)  &  0.78 &  3.14 &  1.57 &     0.59 &  5.88 &  4.12 &  10.39 &  1.18 &  2.35 &  1.76 &  4.12 &  0.78 &  0.20 &  0.20 &  12.75 &   9.41 &  2.16 &   29.41 &   4.51 &  8.63 &      26.86 &  11.18 &  12.75 &  2.35 & 28.43 &  12.35 &  7.25 &  0.39 \\
			Spec.\,(\%) & 56.47 & 50.69 & 54.12 &    42.83 & 47.51 & 51.69 &  51.37 & 49.22 & 46.47 & 56.57 & 42.80 & 55.80 & 44.99 & 51.37 &  44.71 &  60.00 & 43.73 &   23.92 &  48.22 & 32.99 &      48.04 &  43.14 &  47.65 & 72.11 & 34.50 &  71.97 & 60.24 & 46.86 \\
			$R^2$       &  0.88 &  0.99 &  0.99 & -2e3 &  0.99 &  0.99 &  -3.83 &  0.46 &  0.35 &  0.97 & -0.31 &  0.94 &  0.98 &  0.98 & -48.64 & -25.55 &  0.96 & -114.70 &   0.83 &  0.50 & -213e3 & -18.09 & -99.89 & -0.62 & -1.85 &  -3.38 & -5.76 &  0.84 \\
			\bottomrule
	\end{tabular}}
\end{table}

%% file: embedding_visualization.tex
\begin{figure}[h]
    \centering
    \includegraphics[width=.98\linewidth]{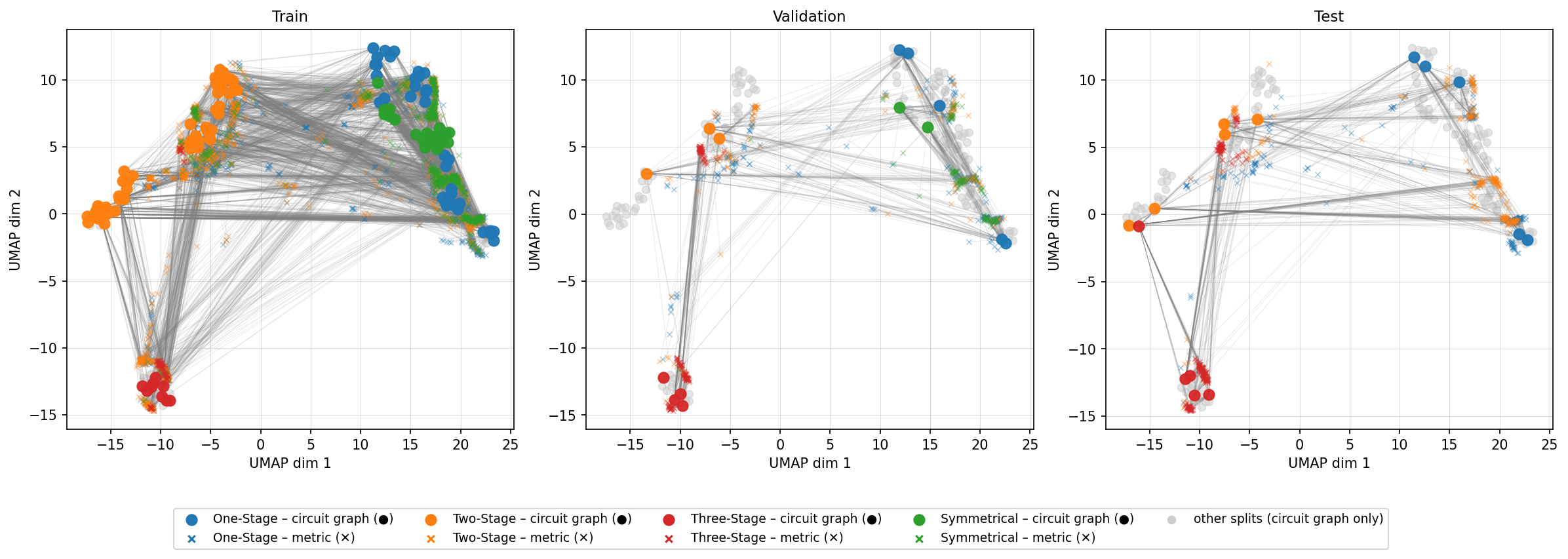}
    \caption{UMAP visualization of the learned embedding space for queries using
    DC gain and F\textsubscript{t}.}
    \label{fig:umap_dc_gain_ft}
\end{figure}

To qualitatively inspect the learned topology-metric embedding space, we
visualize the embeddings using UMAP~\cite{mcinnes2018umap} in Figs.~\ref{fig:umap_dc_gain_ft} using the DC gain and F\textsubscript{t} asa query setting. 
Circuit topology embeddings are shown as circles, while metric embeddings are shown
as crosses. For readability, we subsample the Pareto-optimal metric embeddings shown in the
plots, using 5\% of the available metric vectors per topology while keeping at
least 50 samples whenever available. Colors indicate the topology family, and the three panels separate
the train, validation, and test splits.

%% file: ae.tex
\section{Artifact Appendix}

\subsection{Abstract}

The artifact contains more than 100 1-, 2-, and 3-stage operational amplifier (opamp) topologies in the open-source SkyWater130 PDK, each shipped with structural and functional annotations, topology-specific sizing constraints (symmetry and sizing rules), an Ngspice testbench, and a multi-objective NSGA-II sizing run producing around $10^3$ Pareto-optimal and $10^6$ intermediate sizing-performance pairs per topology. Additionally, a GitHub repository containing Python scripts to help evaluate the proposed 6 benchmarks described in the paper is provided. Thus, the data presented here constitutes the most comprehensive dataset for ML development in over four decades of analog IC design automation. The dataset, ground-truth annotations, and reference implementations of the metric-defining routines are released under an open license, enabling reproducible and directly comparable evaluation across heterogeneous analog design automation methods.

\subsection{Artifact check-list (meta-information)}

{\small
	\begin{itemize}
		\item {\bf Program: } This is a Benchmark
		\item {\bf Compilation: } Only Python + Python-libraries needed
		\item {\bf Model: } User dependent choice, not part of the benchmark
		\item {\bf Data set: } Together with the tasks and metrics, the datasets are the main part of the work. All datasets are included. Size $>100$ GB
		\item {\bf Run-time state: } Some Benchmark metrics are sensitive to the selection of the test set. To this end, test sets are already provided. They were selected randomly from the full dataset.
		\item {\bf Metrics: } For all metrics suggested by the Benchmark, see paper. Code is provided for the evaluation of metrics that are not straight forward like precision or recall are.
		\item {\bf Output: } Output of the Benchmark dataset is mostly the dataset itself, including simulation, etc. (See list in paper introduction). Output expected for the metric evaluation depends on the benchmark.
		\item {\bf Expected results and tolerances: } Application specific.
		\item {\bf How much disk space required (approximately)?: } 110 Gb
		\item {\bf How much time is needed to prepare workflow (approximately)?: } Depending on the task, it can take a few minutes to roughly 20 minutes. The latter in the case that an Ngspice installation is needed for simulating results.
		\item {\bf Publicly available?: } Yes
		\item {\bf Code licenses (if publicly available)?: } CC BY 4.0
		\item {\bf Zenodo DOI for archived artifact: } 10.5281/zenodo.21759759
	\end{itemize}
}

\subsection{Description}

\subsubsection{How to access}

The dataset can be accessed at \url{https://zenodo.org/records/21759759} with Zenodo DOI 10.5281/zenodo.21759759. The GitHub repository that includes the evaluation scripts can be accessed in \url{https://github.com/FilipeAz/LOADBench-Scripts}. The compressed dataset has an approximate space of 40 Gb. Full dataset after decompression is approximately 110 Gb. 

\subsubsection{Software dependencies}

An Ngspice testbench is included with each opamp. To be able to successfully run this testbench, we recommend the Ngspice version 44.2. Regarding the evaluation scripts, we recommend Python version 3.9 or higher. Then, depending on the benchmark, the following python libraries' dependencies exist (other versions might also work, but they were not tested):

\begin{itemize}
	\item \textbf{Circuit Generation}: networkx version 3.2.1;
	
	\item \textbf{Topology Selection}: pandas version 2.3.3;
	
	\item \textbf{Optimization}: numpy version 2.0.2, pandas version 2.3.3, pygmo version 2.19.5;
	
	\item \textbf{Inverse Sizing}: numpy version 2.0.2, pandas version 2.3.3, scikit-learn version 1.6.1.
	\item \textbf{Simulation with NGSpice:} See ngspice website:\\ \url{https://ngspice.sourceforge.io/download.html}
\end{itemize} 

\subsubsection{Commercial software and PDK dependencies}

The dataset uses the open-source SkyWater130 PDK and Ngspice simulator. However, since only a small amount of transistor's models are used in the topologies of the dataset, a reduced SkyWater130 PDK is shipped alongside the dataset.

\subsection{Installation}

As the dataset is quite large, depending on the goal of the user, download the applicable dataset files from the Zenodo link. For example, for the inverse sizing benchmark, only the reduced PDK folder and the \path{TopologySelectionAndInverseSizingBenchmark.zip} are needed. 

For a smoke test of the testbench simulation, download the \path{TopologySelectionAndInverseSizingBenchmark.zip} and \path{reduced_sky130_combined.zip} files. Unzip \path{reduced_sky130_combined.zip} into a folder named \path{reduced_sky130_combined}, and extract the opamps in \texttt{TopologySelectionAnd\allowbreak{}InverseSizingBenchmark.zip} into the same parent directory, so both the opamp folders and the reduced PDK folder are at the same level. Then go into  \path{SimpleThreeStageOpAmp_SimpleOpAmp1_1_4/example_run} and run:

\begin{verbatim}
	$ cd example_run
	$ ngspice CombinedTestBench.cir
	$ python3 standalone_tb_extract.py ./
\end{verbatim}

Access the \path{TB.extract} to see the simulation results. The only requirements are a Ngspice installation and Python installation with the numpy library. Other opamps can be tested but keep in mind that the sizing values in the \path{params.cir} file are random and may lead to simulation errors.

\subsection{Evaluation and expected results}

For the SimpleThreeStageOpAmp\_SimpleOpAmp1\_1\_4, the simulation results should be the following:

\begin{itemize}
	\item P = 0.001447018559781998483688667529
	\item Voffset = 0.0007198916816605338553358706153
	\item icmr\_min = 0.4216533 and icmr\_max = 1.0113
	\item vout\_min = 0.957661 and vout\_max = 1.658198
	\item dc\_gain = 98.47120385038306
	\item ft = 7798252.781202488
	\item phaseMargin = 61.6774692653963
	\item CMRR = 63.336469023814175
	\item PSRR = 63.45675556689272
	\item SR\_pos = 32386929.446193527
	\item SR\_neg = -1078836455.3425071
	\item area = 7516.946499999999
\end{itemize}

\subsection{Experiment customization}

The dataset is easy to use across many applications since all information is stored in standard XML and CSV files.